\documentclass{article}

\usepackage[preprint,nonatbib]{neurips_2024}

\usepackage[utf8]{inputenc} 
\usepackage[T1]{fontenc}    
\usepackage{hyperref}       
\usepackage{url}            
\usepackage{booktabs}       
\usepackage{amsfonts}       
\usepackage{nicefrac}       
\usepackage{microtype}      
\usepackage[table]{xcolor}
\usepackage{listings}
\usepackage{algorithm}
\usepackage{algorithmic}
\usepackage{times}
\usepackage{soul}
\usepackage{url}
\usepackage{graphicx}
\usepackage{amsmath}
\usepackage{amsthm}
\usepackage{algorithm}
\usepackage{algorithmic}
\usepackage{graphicx}
\usepackage{amsmath}
\usepackage{tabularx} 
\usepackage{multirow} 
\usepackage{leftidx}
\usepackage{hyperref}
\usepackage{wrapfig}

\usepackage{enumitem}
\usepackage{tabularx}
\usepackage{xcolor}         
\usepackage{graphicx}
\usepackage{amsmath} 
\usepackage{amsfonts}       
\usepackage{nicefrac}       
\usepackage{microtype}      
\usepackage{booktabs}       
\title{Retrieval Meets Reasoning: Even High-school Textbook Knowledge Benefits Multimodal Reasoning}

\author{%
Cheng Tan$^{1*}$\quad Jingxuan Wei$^{2}$\thanks{Equal contribution.}\quad Linzhuang Sun$^{2}$\quad Zhangyang Gao$^{1}$\quad Siyuan Li$^{1}$\\ 
\textbf{Bihui Yu$^{2}$\quad Ruifeng Guo$^{2}$\quad \ Stan Z. Li}$^{1}$\thanks{Corresponding author.} \\
$^{1}$Westlake University $^{2}$Shenyang Institute of Computing Technology, Chinese Academy of Sciences\\
}

\begin{document}

\maketitle

\begin{abstract}
Large language models equipped with retrieval-augmented generation (RAG) represent a burgeoning field aimed at enhancing answering capabilities by leveraging external knowledge bases. Although the application of RAG with language-only models has been extensively explored, its adaptation into multimodal vision-language models remains nascent. Going beyond mere answer generation, the primary goal of multimodal RAG is to cultivate the models' ability to reason in response to relevant queries. To this end, we introduce a novel multimodal RAG framework named RMR (\textbf{R}etrieval \textbf{M}eets \textbf{R}easoning). The RMR framework employs a bi-modal retrieval module to identify the most relevant question-answer pairs, which then serve as scaffolds for the multimodal reasoning process. This training-free approach not only encourages the model to engage deeply with the reasoning processes inherent in the retrieved content but also facilitates the generation of answers that are precise and richly interpretable. Surprisingly, utilizing solely the ScienceQA dataset, collected from elementary and high school science curricula, RMR significantly boosts the performance of various vision-language models across a spectrum of benchmark datasets, including A-OKVQA, MMBench, and SEED. These outcomes highlight the substantial potential of our multimodal retrieval and reasoning mechanism to improve the reasoning capabilities of vision-language models.
\end{abstract}

\section{Introduction}
While deep learning and its applications have been widely explored in recent years~\cite{tan2021co,gao2022simvp,li2023masked,li2023semireward,tan2023temporal,li2023moganet,tan2022simvp,li2024switch,tan2023openstl}, retrieval-augmented generation (RAG) has rapidly emerged as a cornerstone in the development of large language models (LLMs), enabling them to enhance their capabilities by leveraging external knowledge bases~\cite{yu2022retrieval,jiang2023active,chen2024benchmarking}. Integrating LLMs with RAG has found its most impactful application within language-centric models, where the dynamic interplay between retrieved content and answer generation significantly elevates the quality and relevance of responses~\cite{chang2024survey,zhao2024explainability,chang2024survey}. While early works have demonstrated that incorporating directly retrieved information into language models can improve the quality of the generated content~\cite{izacard2023atlas}, subsequent developments have involved refinement and mitigate the potential noise associated with the raw retrieval results~\cite{xu2023recomp,yu2023chain,asai2023self}, thus ensuring that the content generated is not only accurate but also contextually enriched.

\begin{figure}[ht]
\centering
\vspace{-2mm}
\includegraphics[width=0.98\linewidth]{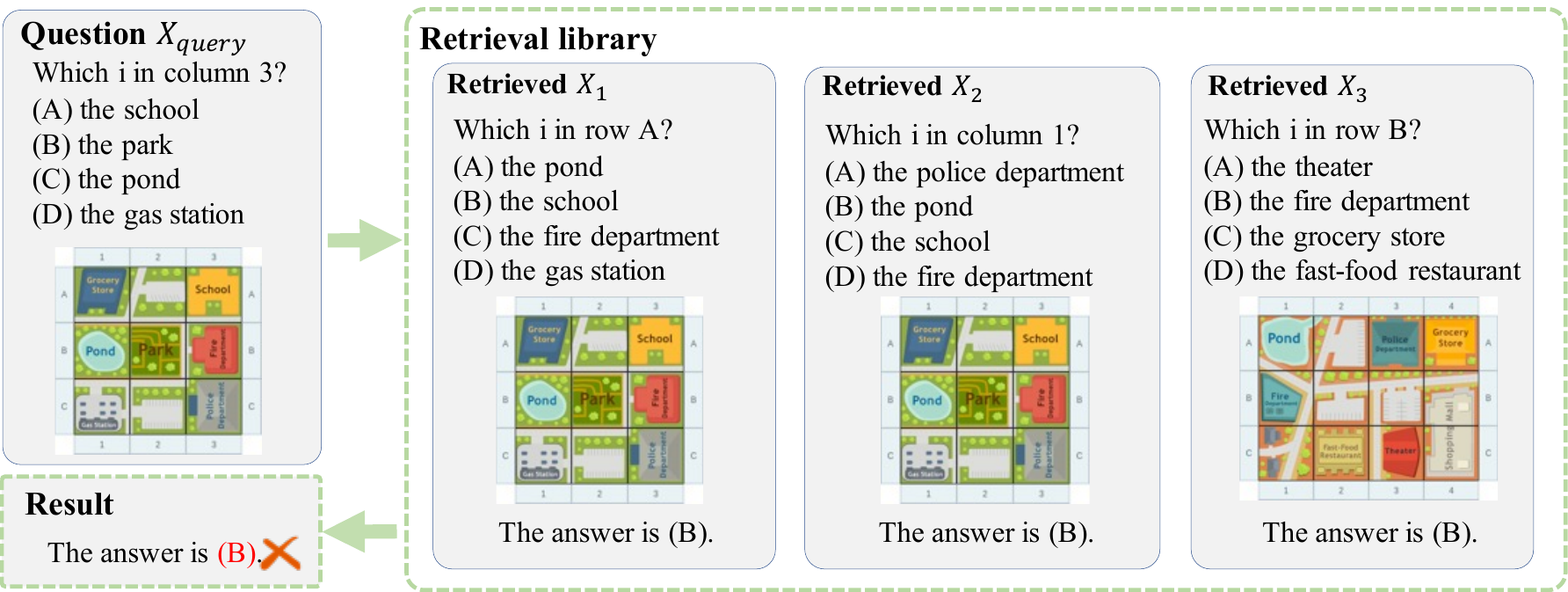}
\vspace{-2mm}
\caption{limitations of multimodal retrieval enhancement with simple question-answer pairs.}
\vspace{-2mm}
\label{fig:limitation_example}
\end{figure}

Although the integration of RAG with large language models has been extensively explored, its adaptation to multimodal scenarios that encompass both visual and textual inputs remains relatively nascent~\cite{chen2022murag,yasunaga2023retrieval,yan2024corrective}. Notably, in the domain of visual question answering (VQA), where queries comprise concise textual prompts paired with complex visual data, the requirements for integration diverge significantly from traditional text-centric approaches~\cite{yan2024corrective,cui2024more}. Traditional RAG systems, initially designed for text-heavy applications, encounter substantial challenges when applied directly to multimodal tasks. In these contexts, textual data often provides insufficient contextual cues, failing to bridge the interpretative demands of rich visual information. This fundamental limitation is critical: models trained predominantly on textual data struggle to effectively capture the nuanced complexity of visual information, leading to significant gaps in the model's ability to accurately interpret and reason about visual content. For instance, as illustrated in Figure~\ref{fig:limitation_example}, even when the three most pertinent pieces of information are retrieved in response to a query, the model may still fail to engage with the underlying reasoning processes embedded within the retrieved content. Instead of synthesizing insights from these inputs, the model may default to merely replicating answers, which can result in inaccuracies. This highlights a critical shortfall in current multimodal RAG systems, i.e., their inability to fully leverage the cognitive reasoning demanded by complex multimodal data.

Building on the foundational understanding that RAG significantly enhances the capabilities of large language models, we hypothesize that the ultimate purpose of multimodal RAG extends beyond merely instructing models to generate direct answers. Instead, our goal is to equip models with the ability to engage in cognitive reasoning, akin to human thought processes when confronted with complex, context-rich questions. This perspective underscores the necessity for multimodal RAG to be inherently flexible and open-ended, facilitating deep contemplation and robust reasoning whether the tasks are unimodal or multimodal, and independent of the data modality being retrieved.

To realize this vision, we develop a comprehensive multimodal RAG framework, named RMR (\textbf{R}etrieval \textbf{M}eets \textbf{R}easoning), which seamlessly integrates multimodal retrieval capabilities with  in-context learning (ICL). This framework begins by employing a bi-modality retrieval module to fetch the most pertinent question-answer pairs, which may be unimodal or multimodal. It then integrates these elements into the model's reasoning process, guiding it through the provided rationales associated with each retrieved item. Following this retrieval phase, the model autonomously learns coherent rationales that reflect a deep and meaningful engagement with the given problem.

Remarkably, even when retrieval is limited to the ScienceQA dataset~\cite{scienceqa}, which covers only elementary and high school science curricula, our Retrieval Meets Reasoning (RMR) framework demonstrates substantial enhancements across a variety of open-source multimodal models. Notably, the LLaVA model registers a +7.66\% improvement, Qwen-VL achieves a +9.93\% increase, InternLM-XComposer2-VL records a +5.33\% improvement, Gemini achieves a +33.67\% increase. Furthermore, when evaluated against diverse datasets such as A-OKVQA~\cite{schwenk2022okvqa}, MMBench~\cite{liu2023mmbench}, and SEED-Bench~\cite{li2023seed}, RMR consistently delivers significant performance gains by leveraging the specialized knowledge embedded in the high-school curriculum from the ScienceQA dataset.

The main contributions of this work are as summarized as follows:
\begin{itemize}[leftmargin=4mm]
\item We introduce a comprehensive multimodal RAG framework, Retrieval Meets Reasoning (RMR), designed to enhance the reasoning capabilities of multimodal models, enabling them to generate answers through cognitive processes rather than merely replicating responses.
\item We develop a bi-modality retrieval module that effectively bridges the gap between unimodal and multimodal data, ensuring robust and accurate retrieval outcomes.
\item Despite its training-free manner, RMR demonstrates significant effectiveness across a variety of multimodal models and datasets, showcasing its capability to improve multimodal reasoning tasks.
\end{itemize}

\section{Related Work}

\paragraph{RAG in LLMs} 
Despite recent advancements in deep learning and large language or vision models~\cite{wei2023enhancing,tan2023boosting,cao2024survey}, retrieval-augmented generation (RAG) improves language model capabilities by integrating external knowledge into the generation process~\cite{li2022survey,gao2023retrieval,cai2022recent,cheng2024lift}. The development of RAG was initiated with the introduction of dense retrievers, which radically transformed how responses are generated by utilizing externally sourced information~\cite{lewis2020retrieval, yu2022retrieval, zhao2024retrieval}. Initial efforts primarily focused on refining the interaction between the retriever and the generator~\cite{lewis2020retrieval,guu2020retrieval}, leading to the production of contextually enriched responses. As a reliable enhancement, RAG has been foundational in allowing Large Language Models (LLMs) to exploit the vast reservoir of knowledge they encompass for various applications, including question answering, dialogue systems, and summarization~\cite{he2021efficient,xu2023recomp,borgeaud2022improving}.

Recent advancements in RAG have been directed at addressing specific challenges, such as reducing hallucination phenomena and integrating outdated and obscure long-tail knowledge. An active retrieval mechanism that adaptively selects the most relevant knowledge pieces is employed to provide up-to-date information for the generation process~\cite{jiang2023active}. Building on this, ActiveRAG~\cite{xu2024activerag} is proposed to incorporate an active learning mechanism that not only retrieves pertinent information but also synthesizes it with existing knowledge, markedly enhancing the model's ability to handle knowledge-intensive tasks by dynamically integrating information. Although the development in RAG has significantly enhanced the synergy between retrieval and generation, the focus has primarily been on text-based applications, with limited exploration into multimodal scenarios.

\paragraph{RAG for ICL} In-context learning (ICL) has revolutionized the functionality of LLMs, enabling them to adapt to new tasks by leveraging a few contextual examples provided directly within their input. This shift towards using retrieved demonstrations to facilitate ICL has increased the flexibility of LLMs across various applications~\cite{dong2022survey,xu2024context,zhang2024makes}. The technique of demonstration retrieval, which involves selecting few-shot examples specifically tailored to the query, not only boosts task performance but also helps to mitigate biases that arise from manual or random selection of demonstrations. A key development has been the optimization of retrieval objectives, which ensures that the demonstrations are both pertinent to the query and diverse enough to offer a comprehensive context~\cite{xu2024context}.

Expanding ICL into multimodal tasks represents a significant advancement, tackling the complex challenge of integrating textual and visual data. The extension of ICL into multimodal tasks represents a significant leap forward, addressing the inherent complexity of integrating textual and non-textual data. MM-Retrieval~\cite{liu2023retrieval} is a concurrent work that introduces a retrieval-augmented multi-modal CoT reasoning approach, which dynamically selects demonstration examples by leveraging cross-modal and intra-modal similarities. However, it operates by employing modality-specific retrievers to gather demonstrations, which are then directly structured into a chain-of-thought format.

\section{Retrieval Meets Reasoning}

\subsection{Preliminaries}

In vision-language tasks, the objective is to develop a mapping $\mathcal{F}_\Theta: \mathcal{X} \rightarrow \mathcal{Y}$, where $\mathcal{X}$ represents multimodal inputs that include both textual and visual elements, and $\mathcal{Y}$ denotes the corresponding outputs. Formally, given a dataset $\mathcal{D} = \{\mathcal{X}, \mathcal{Y}\}$, each input $X \in \mathcal{X}$ can be decomposed into $X = (T, I)$, where $T$ denotes the text component and $I$ denotes the image component. In certain cases, one of them may be absent, i.e., $T=\emptyset$ or $I=\emptyset$, resulting in modality-incomplete inputs.

The primary goal is to accurately predict the output $Y \in \mathcal{Y}$ for a given input $X$. This task can be expressed as identifying the answer $Y$ that maximizes the conditional probability $p(Y \mid T, I)$ given the text component $T$ and the image component $I$:
\begin{equation}
\mathcal{F}_\Theta(X) = \arg\max_{Y'} p(Y' \mid T, I)
\end{equation}
Here, $p(Y' \mid T, I)$ denotes the probability of a candidate answer $Y'$ given the inputs $T$ and $I$. For the supervised learning setting, the model optimal parameters $\Theta^*$ are those that minimize the loss function $\mathcal{L}$, which quantifies the discrepancy between the predicted answer $\mathcal{F}_\Theta(X)$ and the ground truth answer $Y$. This optimization is: $\Theta^* = \arg\min_{\Theta} \mathcal{L}(\mathcal{F}_\Theta(X), Y)$. However, in our approach, we employ an in-context learning strategy based on pre-trained large language models (LLMs). This method uses the retrieved content as contextual information without re-training the model, enabling it to obtain strong reasoning ability effectively.

\subsection{Bi-modality Retrieval Module}

\begin{figure}[ht]
  \centering
  \vspace{2mm}
  \includegraphics[width=\linewidth]{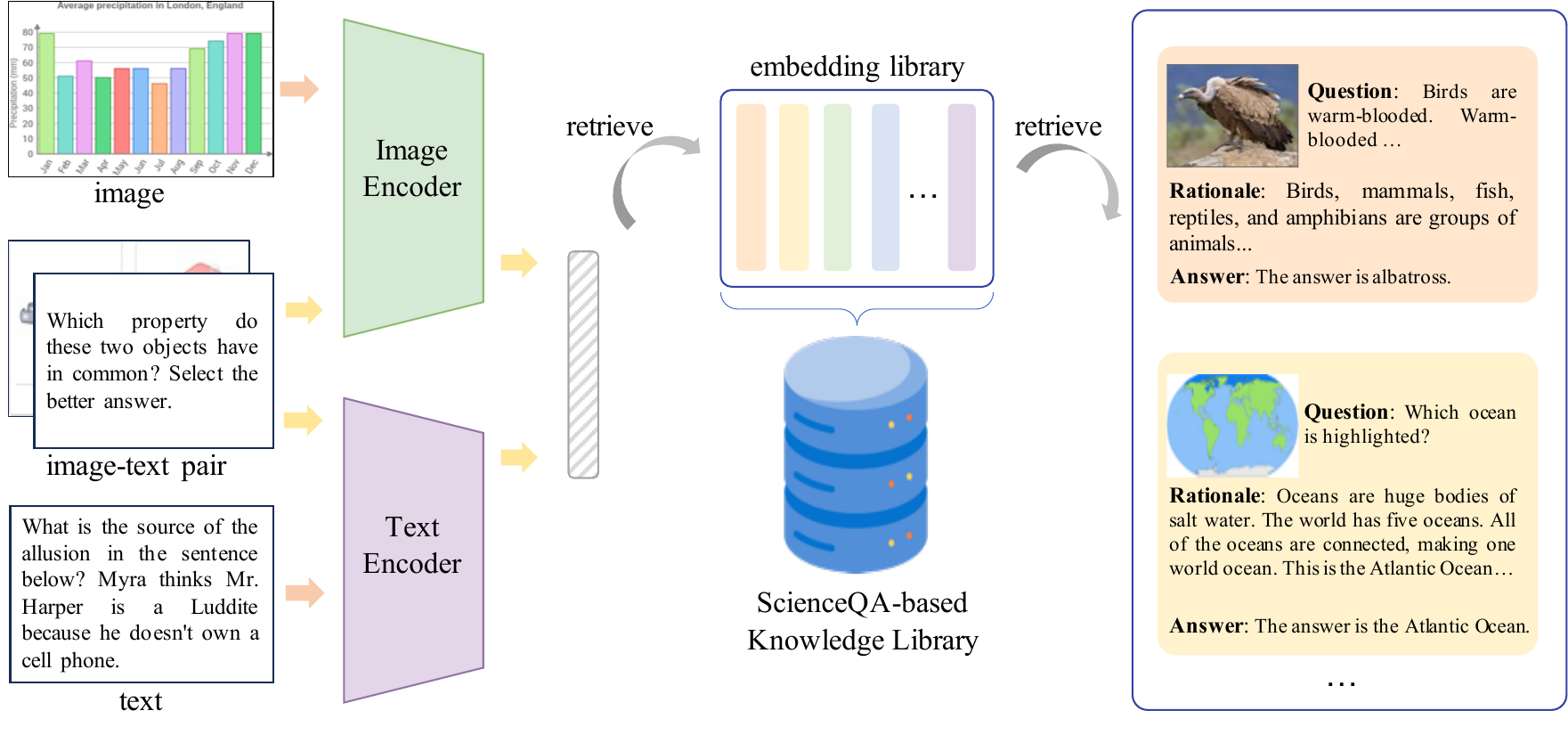}
  \vspace{-2mm}
  \caption{The overall architecture and the retrieval mechanism of the bi-modality retrieval module.}
\label{fig:retrieval_module}
\end{figure}

To unify the retrieval module in multimodal models for both text and visual modality, we propose a bi-modality retrieval module based on the Contrastive Language-Image Pre-training (CLIP)~\cite{radford2021learning} framework, as shown in Figure~\ref{fig:retrieval_module}. This module is designed to handle various cases where the input may consist of complete image-text pairs, image-only inputs, or text-only inputs. The core idea is to create a robust embedding representation that can effectively capture the relevant information across different modalities and retrieve the most pertinent examples to guide the reasoning process.

\paragraph{Embedding Representation} The retrieval module begins by computing embeddings for the inputs using the CLIP model, which is adept at handling both textual and visual data. The embedding strategy is adaptive, accommodating the varying availability of modalities within the input. For image-text pairs $(T, I)$, we calculate the mean pooling of CLIP's text embedding $\boldsymbol{h}_T = \mathrm{CLIP}_T(T) \in \mathbb{R}^{d}$ and image embedding $\boldsymbol{h}_I = \mathrm{CLIP}_I(I) \in \mathbb{R}^{d}$, where $d$ denotes the embedding dimension. Here, $\mathrm{CLIP}_T(\cdot)$ and $\mathrm{CLIP}_I(\cdot)$ denote the text and image encoders of CLIP, respectively. The combined embedding $\boldsymbol{h}_X$ is then computed as the average of these text and image embeddings. For image-only inputs, we use the CLIP image embedding $\boldsymbol{h}_I$, and for text-only inputs, we use the CLIP text embedding $\boldsymbol{h}_T$. Thus, the item embedding $\boldsymbol{h}_X \in \mathbb{R}^{d}$ for any input $X$ is defined as:
\begin{equation}
  \boldsymbol{h}_{X} = \begin{cases} 
  \frac{\boldsymbol{h}_{T} + \boldsymbol{h}_{I}}{2}, & \text{if } T \neq \emptyset \text{ and } I \neq \emptyset \\
  \boldsymbol{h}_{I}, & \text{if } T = \emptyset \text{ and } I \neq \emptyset \\
  \boldsymbol{h}_{T}, & \text{if } T \neq \emptyset \text{ and } I = \emptyset
  \end{cases}
\end{equation}
This approach ensures robust embeddings regardless of the presence or absence of textual and visual components, thereby providing a flexible and consistent representation of multimodal inputs.

\paragraph{High-school Knowledge Library} We construct a comprehensive knowledge embedding library $\mathcal{H} = \{\boldsymbol{h}_X^i\}_{i=1}^N$ using the ScienceQA dataset, which is derived from elementary and high school textbooks. This dataset is particularly valuable because it includes detailed rationales for each answer, forming question-rationale-answer triplets $(Q_i, R_i, A_i)$ for each sample. The structured nature of this dataset provides a rich source of contextual information that is essential for training models to understand and reason about both textual and visual data comprehensively. 

\paragraph{Retrieval Mechanism} The retrieval mechanism employed for identifying relevant triplets operates based on the cosine similarity between the query embedding $\boldsymbol{h}_{X}^{query}$ and each triplet embedding $\boldsymbol{h}_{X}^i$ stored in the library. The cosine similarity is defined as:
\begin{equation}
\text{sim}(\boldsymbol{h}_{X}^{query}, \boldsymbol{h}_{X}^i) = \frac{\boldsymbol{h}_{X}^{query} \cdot \boldsymbol{h}_{X}^i}{|\boldsymbol{h}_{X}^{query}| |\boldsymbol{h}_{X}^i|},
\end{equation}
For each query, the top-$k$ triplets with the highest similarity scores are retrieved. This process ensures that the most contextually relevant examples are selected. The retrieval process can be expressed as:
\begin{equation}
\mathcal{R}(X_{query}) = \{(Q_i, R_i, A_i) \mid i \in \text{Top-}k\left(\text{sim}(\boldsymbol{h}_{X}^{query}, \boldsymbol{h}_{X}^i), \forall i \in \{1,...,N\} \right)\},
\end{equation}
where $\mathcal{R}(X_{query})$ denotes the set of retrieved triplets for the given query $X_{query}$, and $\text{Top-}k(\cdot)$ represents the function that selects the top-$k$ items based on the cosine similarity scores.

\subsection{Learn to Reasoning from the Retrieved Content}

\begin{figure}[ht]
  \centering
  \includegraphics[width=0.9\linewidth]{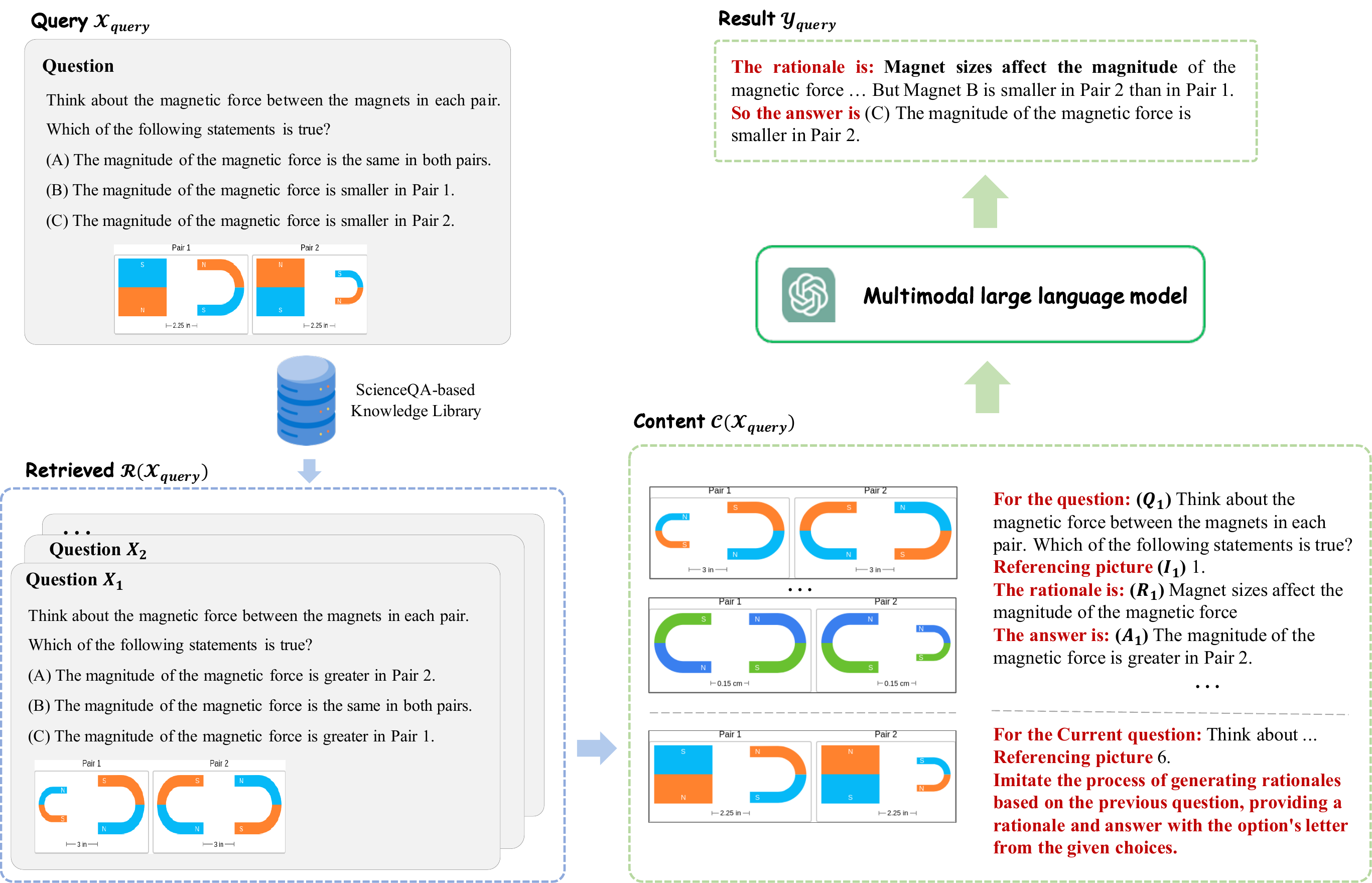}
  \caption{The reasoning process from the retrieved content. The model uses the organized context from retrieved question-rationale-answer triplets to generate answers.}
\label{fig:reasoning_process}
\end{figure}

Given the retrieved question-rationale-answer (QRA) data, we organize them into a structured format to teach the model how to reason, as illustrated in Figure~\ref{fig:reasoning_process}. This section details the process in leveraging the retrieved content to enhance the model's reasoning capabilities.

For a given input $X_{query} = (T_{query}, I_{query})$, suppose we retrieve the top-$k$ relevant QRA triplets $\mathcal{R}(X_{query}) = \{(Q_i, R_i, A_i)\}_{i=1}^k$. These retrieved triplets are used to form a context $\mathcal{C}({X_{query}})$ which provides a structured set of examples to guide the model's reasoning process. The context $\mathcal{C}({X_{query}})$ is constructed as follows:
\begin{equation}
  \mathcal{C}({X_{query}}) = [\mathrm{Example}_1, \mathrm{Example}_2, ..., \mathrm{Example}_k],
\end{equation}
where each example is a concatenation of the question, rationale, and answer:
\begin{equation}
  \mathrm{Example}_i = Q_i \oplus R_i \oplus A_i, \;\; \forall (Q_i, R_i, A_i) \in \mathcal{R}(X_{query}).
\end{equation}
where $\oplus$ denotes the concatenation operation. The context $\mathcal{C}({X_{query}})$ is then used to guide the model's reasoning process, enabling it to learn from the retrieved content and generate accurate and contextually enriched answers. The model is prompted to reason based on the structured examples provided in the context, thereby predicting the answer $\mathcal{Y}_{query}$ given the input $X_{query}$ and the context $\mathcal{C}({X_{query}})$. We define the conditional probability of generating the answer $\mathcal{Y}_{query}$ as:
\begin{equation}
  p(\mathcal{Y}_{query} \mid X_{query}, \mathcal{C}({X_{query}})) = \mathcal{F}_{\Theta}(\mathcal{Y}_{query} \mid X_{query}, \mathcal{C}({X_{query}})),
\end{equation}
where $\mathcal{F}_{\Theta}$ represents the LLMs parameterized by $\Theta$. The final answer $\mathcal{Y}_{query}$ is obtained by:
\begin{equation}
  \mathcal{Y}_{query} = \arg\max_{\mathcal{Y}'} p(\mathcal{Y}' \mid X_{query}, \mathcal{C}({X_{query}})).
\end{equation}

\section{Experiments}

\vspace{-2mm}

\paragraph{Datasets} We employed four multimodal reasoning benchmarks: (i) \textit{ScienceQA}~\cite{scienceqa}, a multimodal question dataset that includes over 21k multiple-choice questions, 3 subjects, 26 topics, 127 categories, and 379 distinct skills. (ii) \textit{A-OKVQA}~\cite{a-okvqa}, a knowledge-based multimodal dataset that includes 25k questions with extensive commonsense and world knowledge. (iii) \textit{MMBench}~\cite{liu2023mmbench}, a dataset comprising 2,974 multiple-choice questions covering 20 ability dimensions. (iv) \textit{SEED-Bench}~\cite{li2023seed}, a large-scale dataset that includes 19k multiple-choice questions, and 12 evaluation dimensions, including both spatial and temporal understanding. We built the high-school knowledge library using the training data from ScienceQA, which employs over 12k question-rationale-answer triplets data.

\vspace{-2mm}

\paragraph{Baselines}
For ScienceQA, we compared our approach against strong baselines across four categories, excluding vision LLMs that specifically fine-tune or train on ScienceQA for a fair comparison: (i) heuristic and expert-guided choices, such as random choice and human evaluation~\cite{scienceqa}; (ii) small multimodal visual question answering models, which include MCAN~\cite{mcan}, Top-Down~\cite{topdown}, BAN~\cite{ban}, DFAF~\cite{dfaf}, ViLT~\cite{vilt}, Patch-TRM~\cite{patchtrm}, and VisualBERT~\cite{visualbert}; (iii) zero-shot instruction-tuned large language models like GPT-3.5~\cite{gpt35} and its CoT-enhanced variants~\cite{scienceqa}, in addition to ChatGPT, GPT-4~\cite{gpt-4}, and Chameleon~\cite{chameleon}; (iv) strong vision LLMs like LLaVA-1.5~\cite{llava}, Qwen-VL~\cite{qwen}, InternLM-XComposer2-VL~\cite{internlm}, and Gemini~\cite{gemini}. Regarding the A-OKVQA dataset, the baselines include state-of-the-art approaches, such as Pythia~\cite{Pythia}, ViLBERT~\cite{vilbert}, LXMERT~\cite{lxmert}, KRISP~\cite{krisp}, GPV-2~\cite{gpv2}, BLIP-2~\cite{blip2}, PICa~\cite{PICA}, IPVR~\cite{IPVR}, PromptCap~\cite{hu2022promptcap}, Prophet~\cite{yu2023prophet}, PaLI-3-VPD~\cite{hu2023visual}, PaLI-X-VPD~\cite{hu2023visual}, and Gemini~\cite{gemini}. For the MMBench~\cite{liu2023mmbench} and SEED-Bench~\cite{li2023seed} datasets, we mainly focus on augmenting Gemini with RMR for the convenient access of the Gemini API.

\vspace{-2mm}
\subsection{Results on ScienceQA}
\vspace{-2mm}

\begin{table}[ht]
\centering
\caption{The comparison on ScienceQA dataset. Question classes: NAT = natural science, SOC = social science, LAN = language science, TXT = text context, IMG = image context, NO = no context, G1-6 = grades 1-6, G7-12 = grades 7-12.}
\small
{\renewcommand\baselinestretch{1.1}
\selectfont
\setlength{\tabcolsep}{0.3mm}{
\begin{tabular}{ccccccccccc}
\toprule
Model & Size & NAT & SOC & LAN & TXT & IMG & NO & G1-6 & G7-12 & AVG \\
\midrule
Random Choice~\cite{scienceqa} & - & 40.28 & 46.13 & 29.25 & 47.75 & 40.08 & 33.66 & 39.35 & 40.67 & 39.83 \\
Human~\cite{scienceqa} & - & 90.23 & 84.97 & 87.48 & 89.60 & 87.50 & 88.10 & 91.59 & 82.42 & 88.40 \\
\hline
MCAN~\cite{mcan} & 95M & 56.08 & 46.23 & 58.09 & 59.43 & 51.17 & 55.40 & 51.65 & 59.72 & 54.54 \\
Top-Down~\cite{topdown} & 70M & 59.50 & 54.33 & 61.82 & 62.90 & 54.88 & 59.79 & 57.27 & 62.16 & 59.02 \\
BAN~\cite{ban} & 112M & 60.88 & 46.57 & 66.64 & 62.61 & 52.60 & 65.51 & 56.83 & 63.94 & 59.37 \\
DFAF~\cite{dfaf} & 74M & 64.03 & 48.82 & 63.55 & 65.88 & 54.49 & 64.11 & 57.12 & 67.17 & 60.72 \\
ViLT~\cite{vilt} & 113M & 60.48 & 63.89 & 60.27 & 63.20 & 61.38 & 57.00 & 60.72 & 61.90 & 61.14 \\
Patch-TRM~\cite{patchtrm} & 90M & 65.19 & 46.79 & 65.55 & 66.96 & 55.28 & 64.95 & 58.04 & 67.50 & 61.42 \\
VisualBERT~\cite{visualbert} & 111M & 59.33 & 69.18 & 61.18 & 62.71 & 62.17 & 58.54 & 62.96 & 59.92 & 61.87 \\
UnifiedQA$_{\text{Base}}$~\cite{UNIFIEDQA} & 223M & 68.16 & 69.18 & 74.91 & 63.78 & 61.38 & 77.84 & 72.98 & 65.00 & 70.12 \\
UnifiedQA$_{\text{Base}}$ w/ CoT~\cite{scienceqa} & 223M & 71.00 & 76.04 & 78.91 & 66.42 & 66.53 & 81.81 & 77.06 & 68.82 & 74.11 \\
\hline
GPT-3.5~\cite{scienceqa} & 173B & 74.64 & 69.74 & 76.00 & 74.44 & 67.28 & 77.42 & 76.80 & 68.89 & 73.97 \\
GPT-3.5 w/ CoT~\cite{scienceqa} & 173B & 75.44 & 70.87 & 78.09 & 74.68 & 67.43 & 79.93 & 78.23 & 69.68 & 75.17 \\
ChatGPT w/ CoT~\cite{gpt-4} & - & 78.82 & 70.98 & 83.18 & 77.37 & 67.92 & 86.13 & 80.72 & 74.03 & 78.31 \\
GPT-4 w/ CoT~\cite{gpt-4} & - & 85.48 & 72.44 & \textbf{90.27} & 82.65 & 71.49 & 92.89 & 86.66 & 79.04 & 83.99 \\
Chameleon + ChatGPT~\cite{chameleon} & - & 81.62 & 70.64 & 84.00 & 79.77 & 70.80 & 86.62 & 81.86 & 76.53 & 79.93 \\
Chameleon + GPT-4~\cite{chameleon} & - & 89.83 & 74.13 & 89.82 & 88.27 & 77.64 & 92.13 & 88.03 & 83.72 & 86.54 \\
\hline
LLaVA-1.5~\cite{llava} & 13B & 70.12 & 76.72&67.64&70.48&71.89&68.92&76.06&61.5&70.86 \\
\rowcolor[HTML]{E7ECE4} LLaVA-1.5+RMR & 13B & 78.11&84.25&74.73&79.81&78.33&74.36&82.05&72.18&78.52 \\
\rowcolor[HTML]{E7ECE4} \textit{+Improvement} & - & +\textit{7.99}&+\textit{7.53}&+\textit{7.09}&+\textit{9.33}&+\textit{6.44}&+\textit{5.44}&+\textit{5.99}&+\textit{10.68}&+\textit{7.66} \\
\hline
Qwen-VL~\cite{qwen} & / & 67.01&66.37&59.36&68.52&67.03&57.00&69.75&56.16&64.89 \\
\rowcolor[HTML]{E7ECE4} Qwen-VL+RMR & / & 70.07&86.16&75.36&70.77&76.50&72.68&78.67&67.90&74.82 \\
\rowcolor[HTML]{E7ECE4} \textit{+Improvement} & / & +\textit{3.06} & +\textit{19.79} & +\textit{16.00} & +\textit{2.25} & +\textit{9.47} & +\textit{15.68} & +\textit{8.92} & +\textit{11.74} & +\textit{9.93} \\
\hline
InternLM-XComposer2-VL~\cite{internlm} & / & 88.19&93.48&78.64&88.47&89.14&80.77&88.66&83.52&86.82\\
\rowcolor[HTML]{E7ECE4} InternLM-XComposer2-VL+RMR & / & \textbf{94.85}&\textbf{97.19}&82.55&\textbf{95.11}&\textbf{96.03}&\underline{85.23}&\textbf{93.98}&\underline{88.86}&\textbf{92.15}\\
\rowcolor[HTML]{E7ECE4} \textit{+Improvement} & / & +\textit{6.66} & +\textit{3.71} & +\textit{3.91} & +\textit{6.64} & +\textit{6.89} & +\textit{4.46} & +\textit{5.32} & +\textit{5.34} & +\textit{5.33}\\
\hline
Gemini~\cite{gemini} & / & 59.68&74.24&41.73&57.72&64.01&47.46&65.20&45.29&58.08\\
\rowcolor[HTML]{E7ECE4} Gemini+RMR & / & \underline{91.79}&\underline{94.26}&\underline{89.64}&\underline{91.40}&\underline{89.69}&\textbf{91.01}&\underline{92.84}&\textbf{89.78}&\underline{91.75}\\
\rowcolor[HTML]{E7ECE4} \textit{+Improvement} & / & +\textit{32.11} & +\textit{20.02} & +\textit{47.91} & +\textit{33.68} & +\textit{25.68} & +\textit{43.55} & +\textit{27.64} & +\textit{44.49} & +\textit{33.67}\\
\bottomrule
\end{tabular}
} 
\par}
\vspace{-2mm}
\label{tab:scienceqa}
\end{table}

We set the number of retrieved examples $k$ to 3 by default. Table~\ref{tab:scienceqa} illustrates the performance of RMR compared to various strong baselines across the ScienceQA dataset. Notably, our approach demonstrates significant improvements in reasoning capabilities without the need for fine-tuning or re-training on the target dataset, adhering to a zero-shot training setting. Despite the relatively low baseline performance of the vanilla Gemini model, which achieved an average accuracy of 58.08\%, the integration of RMR remarkably enhances its reasoning capabilities, achieving an impressive average accuracy of 91.75\%. This represents a substantial improvement of +33.67\% over the baseline. Similarly, for the robust InternLM-XComposer2-VL model, our RMR framework yields a notable improvement of +5.33\%, showcasing its effectiveness in boosting the reasoning capabilities of strong multimodal models. These results underscore the effectiveness of our proposed RMR framework in enhancing the performance of vision-language models across various question classes and contexts, demonstrating its potential to significantly advance the field of multimodal reasoning.

\subsection{Results on A-OKVQA}

Table~\ref{tab:aokvqa_val} presents the performance comparison for the direct-answer task on the A-OKVQA dataset. The table includes a variety of vision-language model combinations across different architectures. The models range from earlier architectures like Pythia and ViLBERT, to more recent and powerful systems such as BLIP-2, PaLM-CoT, and PaLI-X-VPD. Our RMR framework, when integrated with the Gemini model, demonstrates a substantial improvement in performance. The baseline Gemini model achieves a direct-answer accuracy of 44.2\%, but with the incorporation of RMR, this accuracy increases significantly to 63.1\%. This represents an impressive improvement of +18.9\%. 

This improvement highlights the capability of the RMR framework to effectively augment the reasoning and answering performance of existing multimodal models, showcasing its potential to set new benchmarks in visual question answering. The results indicate that RMR not only competes with but also surpasses state-of-the-art methods across diverse model architectures and parameter sizes.

\begin{table*}[ht]
\centering
\vspace{-2mm}
\caption{The comparison of the direct-answer task on the A-OKVQA dataset.}
\small
{\renewcommand\baselinestretch{1.1}
\selectfont
\setlength{\tabcolsep}{1.6mm}{
\begin{tabular}{cccccc}
\toprule
Model & Vision Model & Text Model & Parameters & Direct-answer \\
\midrule
Pythia~\cite{Pythia} & ResNet~\cite{resnet} & BERT~\cite{bert} & 70M & 25.2 \\
ViLBERT~\cite{vilbert} & Faster R-CNN~\cite{fastrcnn} & BERT~\cite{bert} & 300M & 30.6 \\
LXMERT~\cite{lxmert} & Transformer~\cite{attention} & Transformer~\cite{attention} & 220M & 30.7 \\
KRISP~\cite{krisp} & Faster R-CNN~\cite{fastrcnn} & BERT~\cite{bert} & 200M & 33.7 \\
GPV-2~\cite{gpv2} & VinVL~\cite{vinvl} & T5-Base~\cite{t5} & 300M & 48.6 \\
\hline
BLIP-2~\cite{blip2} & CLIP-VIT-LARGE~\cite{radford2021learning} & FlanT5XXL~\cite{FlanT5XXL} & 11B & 53.2 \\
PaLM-CoT~\cite{cot} & - & PaLM~\cite{palm} & 540B &41.5 \\
PICa~\cite{PICA}& VinVL~\cite{vinvl} & GPT-3~\cite{gpt3} & 175B & 42.4 \\
IPVR~\cite{IPVR}& Faster-RCNN~\cite{fastrcnn} & OPT~\cite{opt} & 66B  & 46.4 \\
PromptCap~\cite{hu2022promptcap} & Ofa~\cite{wang2022ofa} & GPT-3~\cite{gpt3} & 175B & 56.3 \\
Prophet~\cite{yu2023prophet} & MCAN~\cite{yu2019deep} & GPT-3~\cite{gpt3}  & 175B & 58.2 \\
PaLI-3-VPD generalist~\cite{hu2023visual} & SigLIP~\cite{zhai2023sigmoid} & UL2~\cite{tay2022ul2} & 5B & 56.5 \\ 
PaLI-X-VPD generalist~\cite{hu2023visual} & ViT~\cite{dehghani2023scaling} & UL2~\cite{tay2022ul2} & 55B & 62.7 \\
\hline
Gemini~\cite{gemini} & - & - & - & 44.2 \\
\rowcolor[HTML]{E7ECE4} Gemini+RMR & - & - & - & \textbf{63.1} \\
\rowcolor[HTML]{E7ECE4} \textit{+Improvement} & - & - & - & +\textit{18.9} \\
\bottomrule
\end{tabular}
} 
\par}
\vspace{-4mm}
\label{tab:aokvqa_val}
\end{table*}

\subsection{Results on MMBench and SEED-Bench}

\begin{figure}[ht]
  \centering
  \includegraphics[width=1.0\linewidth]{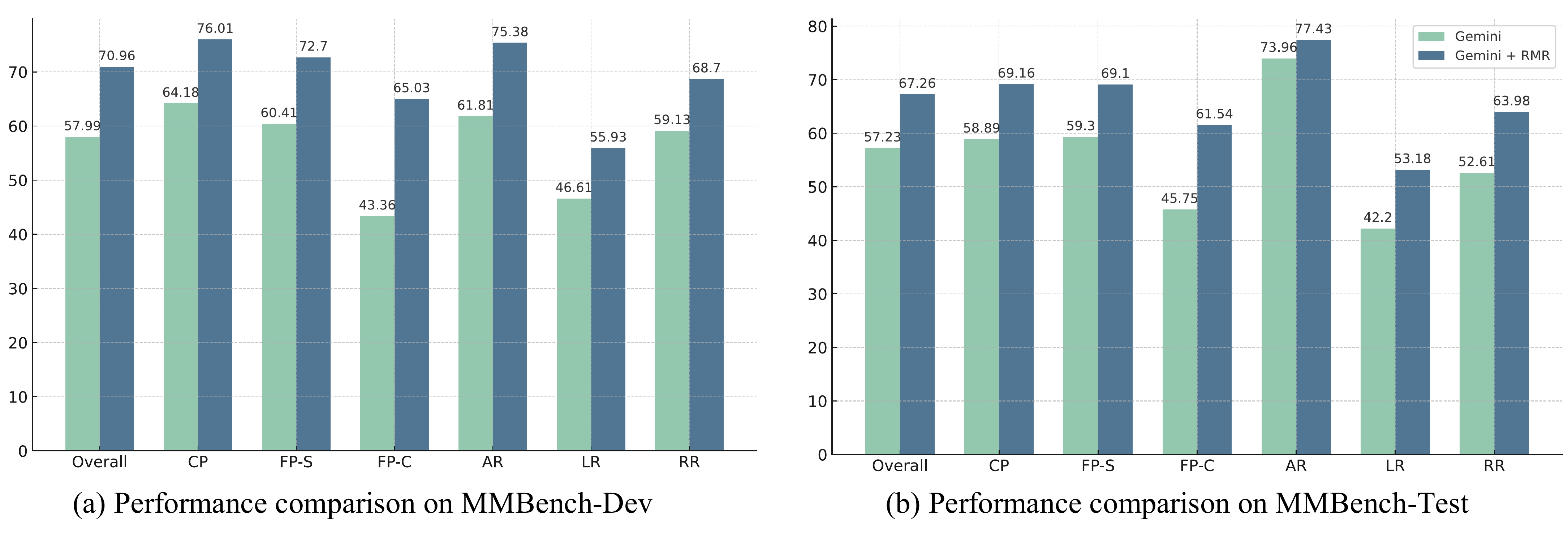}
  \vspace{-4mm}
  \caption{Comparative performance of Gemini and Gemini+RMR on the MMBench-Dev and MMBench-Test datasets.}
\label{fig:mmbench}
\end{figure}

Figure~\ref{fig:mmbench} illustrates the comparative performance of the Gemini model and its enhanced version, Gemini+RMR, on the MMBench-Dev and MMBench-Test datasets across various categories. The categories evaluated include overall performance, conceptual problems (CP), fine-grained perception of cross instances (FP-C), fine-grained perception of single instances, attribute reasoning (AR), logical reasoning (LR), and relation reasoning (RR). For the MMBench-Dev dataset, Gemini+RMR shows marked improvements in all categories compared to the baseline Gemini model. The overall performance of Gemini+RMR is 70.96\%, compared to 57.9\% for Gemini, representing a substantial improvement. Specifically, Gemini+RMR scores 76.01\% in CP, 72.70\% in FP-S, 65.03\% in FP-C, 75.38\% in AR, 55.93\% in LR, and 68.70\% in RR. These results indicate that the RMR framework significantly enhances the model's capability to handle a wide range of multimodal reasoning tasks.

Similarly, on the MMBench-Test dataset, Gemini+RMR outperforms the baseline Gemini model across all categories. The overall performance of Gemini+RMR is 67.26\%, compared to 57.23\% for Gemini. The specific category improvements are as follows: 69.16\% in CP, 69.10\% in FP-S, 61.54\% in FP-C, 77.43\% in AR, 53.18\% in LR, and 63.98\% in RR. These improvements further demonstrate the effectiveness of the RMR framework in enhancing the model's reasoning capabilities across diverse evaluation metrics. Overall, the integration of the RMR framework with the Gemini model leads to significant performance improvements in both the MMBench-Dev and MMBench-Test datasets, underscoring the efficacy of our approach in advancing multimodal reasoning performance.

\begin{wrapfigure}{r}[0cm]{0pt}
  \centering
  \includegraphics[width=0.5\textwidth]{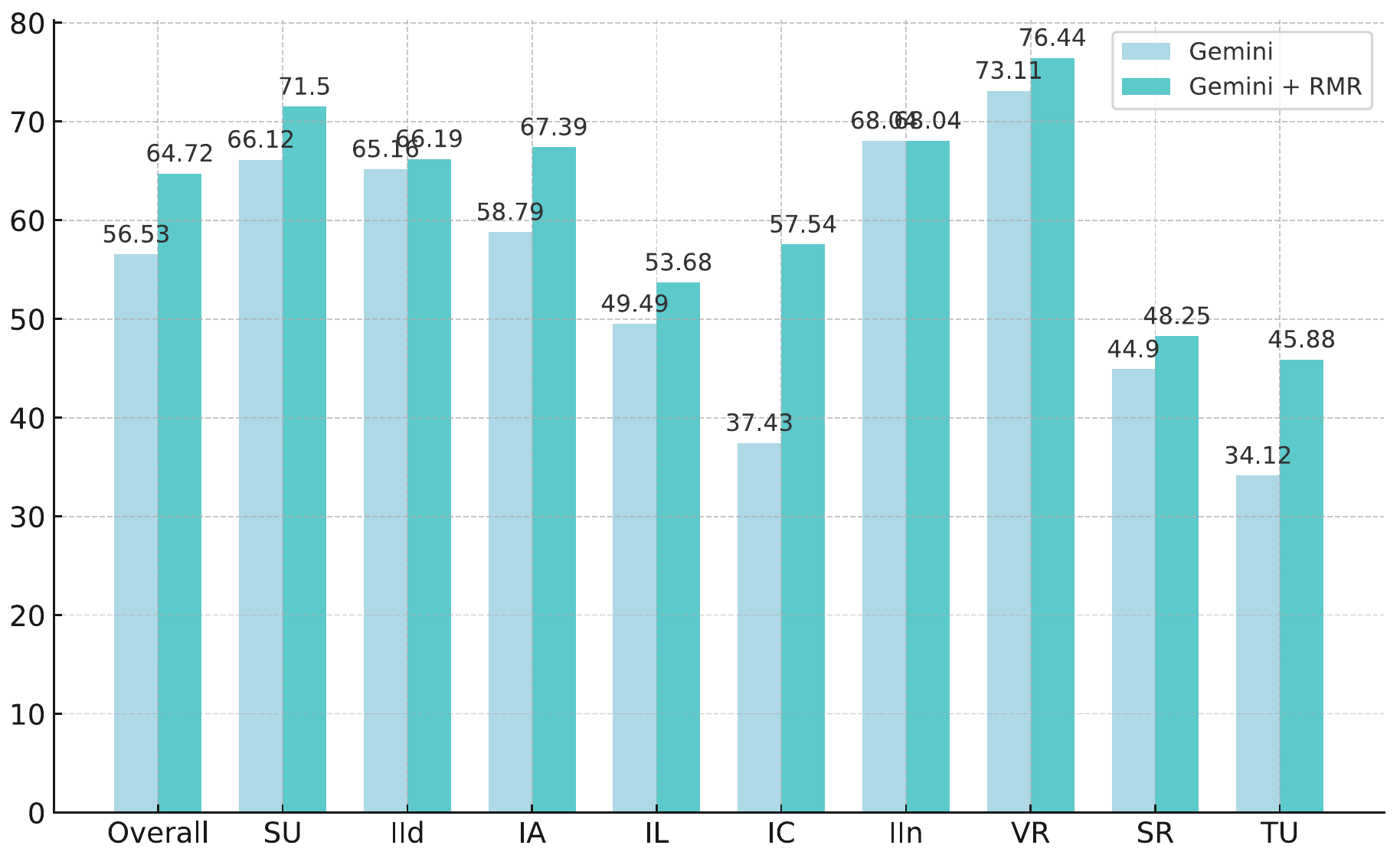}
  \caption{Performance on SEED-Bench dataset.}
  \label{fig:seed-bench}
\end{wrapfigure}

Figure~\ref{fig:seed-bench} presents the comparative performance of the Gemini model and its enhanced version, Gemini+RMR, across various evaluation dimensions on the SEED-Bench dataset. The evaluation categories include overall performance, scene understanding (SU), instance identity (IId), instance attributes (IA), instance location (IL), instance counting (IC), instance interaction (IIn), visual reasoning (VR), spatial relation (SR), and text understanding (TU). The results show significant performance improvements across all categories when integrating the RMR framework with the Gemini model. The overall performance of Gemini+RMR is 64.72\%, compared to 56.53\% for Gemini.

\subsection{Ablation Study}

\paragraph{Effect of Retrieval Size} We investigate the impact of varying the retrieval size on the performance of the RMR framework. The retrieval size $|\mathcal{R}(\mathcal{X}_{query})|$ is defined as the number of question-rationale-answer (QRA) triplets retrieved for each query $\mathcal{X}_{query}$. Table~\ref{tab:ablation_retrieval_size} presents the results of our ablation study on the ScienceQA dataset, showing the performance of Gemini with different retrieval sizes.
\begin{table}[ht]
\centering
\caption{The impact of retrieval size on the performance of RMR on ScienceQA.}
\small
{\renewcommand\baselinestretch{1.1}
\selectfont
\setlength{\tabcolsep}{1.9mm}{
\begin{tabular}{ccccccccccc}
\toprule
Model & NAT & SOC & LAN & TXT & IMG & NO & G1-6 & G7-12 & AVG \\
\midrule
Gemini w/o retrieval & 59.68&74.24&41.73&57.72&64.01&47.46&65.20&45.29&58.08\\
\hline
Gemini w/ $|\mathcal{R}(\mathcal{X}_{query})|=1$ & 88.41&89.43&84.82&87.34&84.53&87.11&89.72&84.05&87.69 \\
Gemini w/ $|\mathcal{R}(\mathcal{X}_{query})|=2$ & 87.79&91.00&84.55&86.75&84.93&86.90&88.88&85.37&87.62\\
Gemini w/ $|\mathcal{R}(\mathcal{X}_{query})|=3$ & {91.79}&{94.26}&{89.64}&{91.40}&{89.69}&{91.01}&{92.84}&{89.78}&{91.75}\\
Gemini w/ $|\mathcal{R}(\mathcal{X}_{query})|=4$ &88.54&92.80&84.27&88.03&86.37&86.41&89.76&85.76&88.33\\
Gemini w/ $|\mathcal{R}(\mathcal{X}_{query})|=5$ & 87.83&89.09&84.45&86.46&83.59&86.69&89.24&83.59&87.22\\
\bottomrule
\end{tabular}
} 
\par}
\label{tab:ablation_retrieval_size}
\end{table}

The results indicate that the optimal retrieval size is $|\mathcal{R}(\mathcal{X}_{query})|=3$, which achieves the highest average accuracy of 91.75\%. Increasing the retrieval size beyond this value leads to a slight decrease in performance, which may be attributed to (i) the inclusion of less relevant examples that could potentially introduce noise into the reasoning process and (ii) the longer context length that may hinder the model's ability to effectively reason over the retrieved content.

\paragraph{Data modality} We compare the performance of the RMR framework using different data modalities: all available data (denoted as "-All"), only text-image pairs (denoted as "-T\&I"), and only text data (denoted as "-T"). Table~\ref{tab:ablation_modality} presents the results of this ablation study on ScienceQA. RMR framework consistently enhances the performance of the models regardless of the data modality used, which can be attributed to the universality of its bi-modality retrieval module.

\begin{table}[ht]
\centering
\caption{The impact of single vs. multi-modal retrieval on ScienceQA.}
\small
{\renewcommand\baselinestretch{1.1}
\selectfont
\setlength{\tabcolsep}{2.7mm}{
\begin{tabular}{ccccccccccc}
\toprule
Model & NAT & SOC & LAN & TXT & IMG & NO & G1-6 & G7-12 & AVG \\
\midrule
LLaVA-All & 70.12&76.72&67.64&70.48&71.89&68.92&76.06&61.5&70.86\\
\rowcolor[HTML]{E7ECE4} w/ RMR& 78.11&84.25&74.73&79.81&78.33&74.36&82.05&72.18&78.52\\
LLaVA-T\&I & 69.56&74.74&86.36&69.69&71.89&-&78.03&56.97&71.89\\
\rowcolor[HTML]{E7ECE4} w/ RMR & 75.19&82.33&95.45&75.58&78.33&-&82.58&68.03&78.33\\
LLaVA-T & 70.76& 88.80&66.86&71.74&-&68.92&73.90&64.37&69.92\\
\rowcolor[HTML]{E7ECE4} w/ RMR & 81.50	& 96.00&73.86&86.57& -&74.36&81.47&74.81&78.69\\
\hline
Qwen-VL-All & 67.01&66.37&59.36&68.52&67.03&57.00&69.75&56.16&64.89\\
\rowcolor[HTML]{E7ECE4} w/ RMR & 70.07&86.16&75.36&70.77&76.50&72.68&78.67&67.90&74.82\\
Qwen-VL-T\&I & 67.25&66.10&77.27&65.23&67.03&-&73.06&52.38&67.03\\
\rowcolor[HTML]{E7ECE4} w/ RMR & 70.31&85.47&90.91&68.50&76.50&-&82.44&62.07&76.50\\
Qwen-VL-T & 66.73&68.00&58.62&73.76&-&57.00&66.10&58.56&62.95\\
\rowcolor[HTML]{E7ECE4} w/ RMR & 69.80&90.40&74.72&74.40&-&72.68&74.52&71.58&73.29\\
\hline
InternLM-All & 88.19&93.48&78.64&88.47&89.14&80.77&88.66&83.52&86.82\\
\rowcolor[HTML]{E7ECE4} w/ RMR & 94.85&97.19&82.55&95.11&96.03&85.23&93.98&88.86&92.15\\
InternLM-T\&I & 85.77&93.98&97.73&86.32&89.14&-&90.83&85.03&89.14\\
\rowcolor[HTML]{E7ECE4} w/ RMR & 95.29&97.25&95.45&95.31&96.03&-&96.78&94.22&96.03\\
InternLM-T & 90.99&90.40&77.84&91.89&-&	80.77&86.25&82.56&84.71\\
\rowcolor[HTML]{E7ECE4} w/ RMR & 94.34&96.80&82.01&94.8&-&85.23&90.89&85.47&88.62\\
\hline
Gemini-All & 59.68&74.24&41.73&57.72&64.01&47.46&65.20&45.29&58.08\\
\rowcolor[HTML]{E7ECE4} w/ RMR &91.79&94.26&89.64&91.40&	89.69&91.01&92.84&89.78&91.75\\
Gemini-T\&I & 55.75&76.83&68.18&54.89&64.01&-&69.91&49.66&64.01\\
\rowcolor[HTML]{E7ECE4} w/ RMR & 86.85&93.72&97.73&87.03&89.69&-&91.18&86.05&89.69\\
Gemini-T & 64.24&58.40&40.62&62.23&-&47.46&60.00&	42.52&52.70\\
\rowcolor[HTML]{E7ECE4} w/ RMR & 97.51&97.60&	89.30&98.35&-	&91.01&94.67&92.14&93.62\\
\bottomrule
\end{tabular}
} 
\par}
\label{tab:ablation_modality}
\end{table}


\section{Conclusion and Limitation}

In this work, we introduce RMR, a multimodal RAG framework designed to enhance the reasoning capabilities of vision LLMs. By leveraging a bi-modality retrieval module, RMR retrieves the most relevant question-rationale-answer triplets from a high-school knowledge library constructed using the ScienceQA dataset. The retrieved triplets are then utilized to form a structured context that guides the model's reasoning process. Extensive experiments on multiple multimodal reasoning benchmarks demonstrate that RMR significantly improves the performance of various vision LLMs.

Despite the promising results, our work has several limitations that warrant further investigation. First, while RMR operates in a training-free manner, which offers convenience and effectiveness, developing a trainable multimodal RAG model could potentially further enhance the reasoning capabilities of vision-language models by allowing the model to adapt more precisely to specific datasets and tasks. Additionally, the high-school knowledge library constructed by ScienceQA may not be comprehensive enough to cover all scenarios, especially for domain-specific questions. 

\bibliography{ref}
\bibliographystyle{plain}

\newpage
\appendix

\section{Qualitative Analysis}

\begin{figure}[ht]
  \centering
  \vspace{-2mm}
  \includegraphics[width=0.90\linewidth]{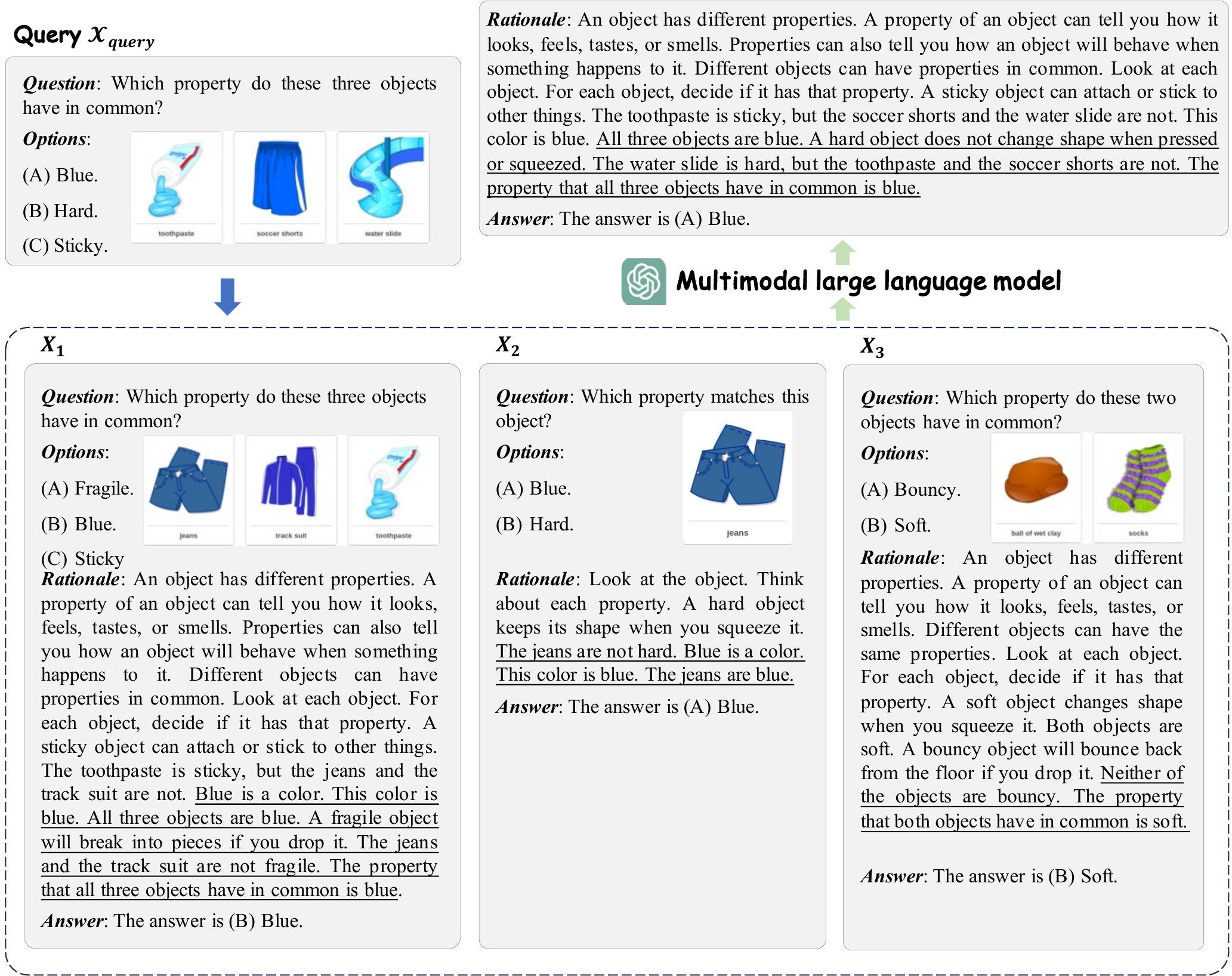}
  \vspace{-2mm}
  \caption{The retrieved data of an image-text pair example.}
  \vspace{-2mm}
\label{fig:example_image1}
\end{figure}

\begin{figure}[h!]
  \centering
  \includegraphics[width=0.90\linewidth]{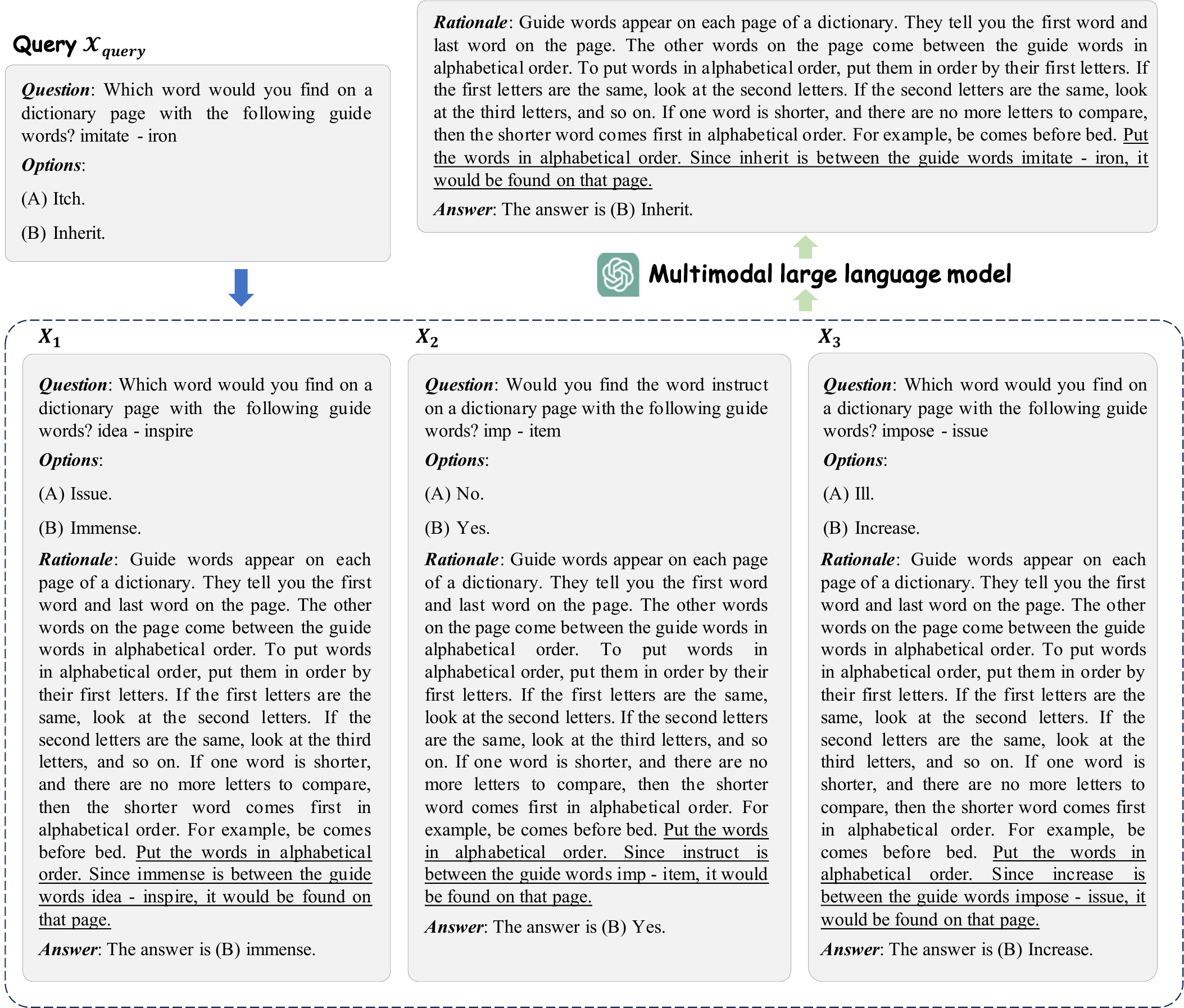}
  \vspace{-2mm}
  \caption{The retrieved data of a text-only example.}
  \vspace{-4mm}
\label{fig:example_text1}
\end{figure}

\begin{figure}[ht]
  \centering
  \vspace{-2mm}
  \includegraphics[width=0.92\linewidth]{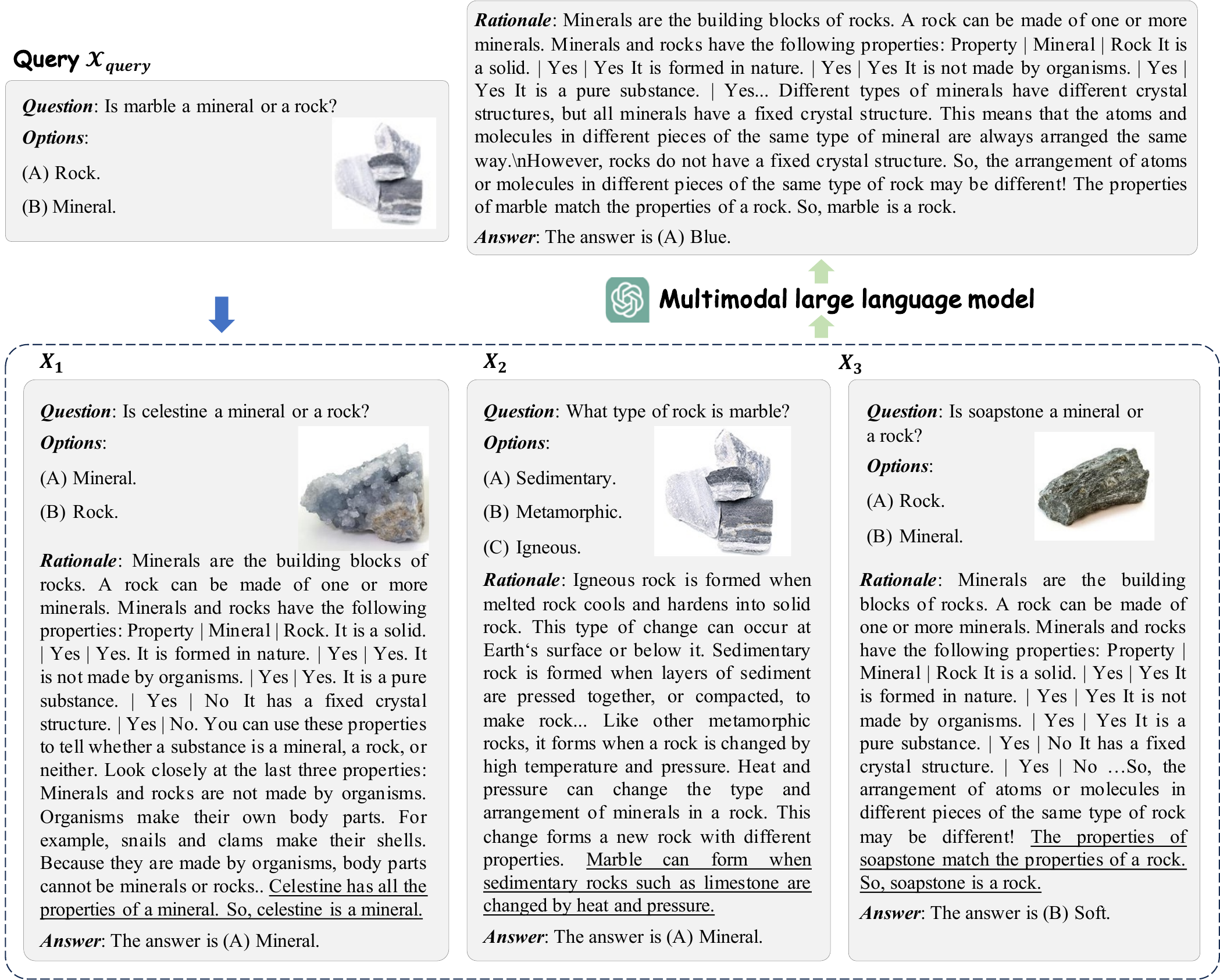}
  \vspace{-2mm}
  \caption{The retrieved data of an image-text pair example.}
\label{fig:example_image2}
\end{figure}

\begin{figure}[!h]
  \centering
  \includegraphics[width=0.92\linewidth]{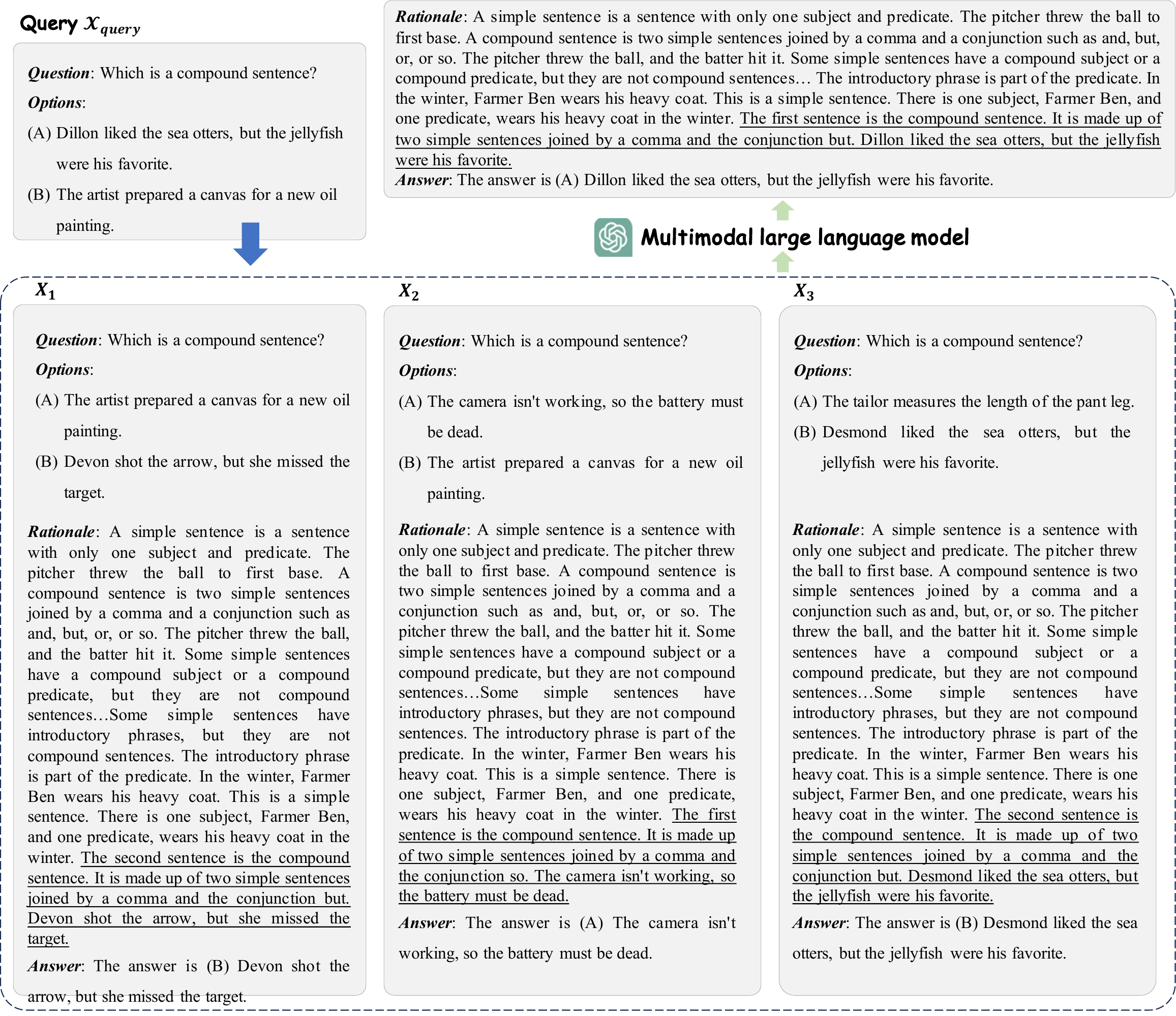}
  \vspace{-2mm}
  \caption{The retrieved data of a text-only example.}
  \vspace{-6mm}
\label{fig:example_text2}
\end{figure}

We present qualitative examples to illustrate the effectiveness of the RMR framework. Figure~\ref{fig:example_image1} showcases an example where the model is asked, "Which property do these three objects have in common?" The retrieved data plays a crucial role in guiding the model's reasoning process. $X_1$ provides a similar question involving different objects. The similarity in the question structure helps the model understand the type of reasoning required, offering a blueprint for approaching the problem. By seeing how the retrieved example addresses the comparison of different objects, the model can apply a similar strategy to the current query. $X_2$ includes a related object, enhancing the model's comprehension of various object properties. The additional context provided by $X_2$ helps the model draw connections between the current objects and previously encountered ones, broadening its understanding and improving its ability to identify common properties. Similar to $X_1$, $X_3$ also presents a question about the common properties of different objects, albeit with a wider variety. The diversity in examples reinforces the model's reasoning skills, helping it generalize its reasoning ability across different contexts and object sets.

Figure~\ref{fig:example_text1} depicts a text-only example where the model is asked, "Which word would you find on a dictionary page with the following guide words?". Three similar questions are retrieved with different words, providing the model with a diverse set of examples to learn from. The retrieved examples help the model understand the structure of the question and the type of reasoning required to answer it. By observing the different words in the retrieved examples, the model can learn to identify the commonalities between the guide words and the target words, enhancing its reasoning capabilities.

Figure~\ref{fig:example_image2} presents an image and text example where the model is asked, "Is marble a mineral or a rock?". The retrieved data provides relevant context that helps the model formulate a more accurate and informed response. The retrieved data $X_1$ contains the question, "Is celestine a mineral or a rock?" This question is analogous to the query about marble, offering the model a direct parallel that aids in understanding how to distinguish between minerals and rocks. The similar structure helps the model apply the reasoning used for celestine to marble, reinforcing the process of categorizing geological substances. $X_2$ poses the more specific question, "What type of rock is marble?" It provides the model with additional information about the classification of marble within the broader category of rocks. This specific context helps the model not only affirm that marble is a rock but also understand its specific type, thereby enriching the model's geological knowledge base. $X_3$ includes the question, "Is soapstone a mineral or a rock?" Similar to $X_1$, it presents another instance of the mineral vs. rock distinction. The inclusion of different substances like soapstone reinforces the model's ability to generalize the reasoning process across various materials, ensuring a robust understanding of the mineral and rock classification.

Figure~\ref{fig:example_text2} presents a text-only example where the model is asked, "Which is a compound sentence?" The retrieved examples all ask the same question but with different options, which provide the model with varied contexts and sentence structures to learn from. The consistent retrieval of questions on compound sentences helps the model understand the syntactic characteristics that define compound sentences. By comparing and analyzing the different options presented in the retrieved examples, the model can improve its ability to identify compound sentences accurately.

\vspace{-2mm}
\section{Broader Impact}
\vspace{-2mm}

Here, we outline several broader impacts of our work:

\paragraph{Hallucination} Similar to other large language models, RMR-integrated models like LLaVA might generate outputs that are not grounded in factual information or the input data provided. This phenomenon, known as hallucination, may raise concerns. Ensuring the reliability and accuracy of outputs in such sensitive domains is crucial, and further research is needed to mitigate these risks.

\paragraph{Biases} The RMR framework inherits biases from its base models, including the vision encoder (CLIP) and the vision LLMs (LLaVA, Qwen-VL, InternLM, and Gemini). These biases may be reflected in the retrieved examples and the model's reasoning process. It is essential to address these biases to ensure fair and unbiased reasoning capabilities.

\paragraph{Energy Consumption} Although our approach is training-free, running inference still incurs energy consumption. The computational resources required for processing large volumes of data. Efforts should be made to optimize inference efficiency and explore sustainable computing practices to minimize the energy footprint of using RMR-enhanced models.

\end{document}